\documentclass{article}

\usepackage{LoD-VGGT}
\usepackage[utf8]{inputenc} 
\usepackage[T1]{fontenc}    
\usepackage{hyperref}       
\usepackage{url}            
\usepackage{booktabs}       
\usepackage{amsfonts}       
\usepackage{nicefrac}       
\usepackage{microtype}      
\usepackage{lipsum}
\usepackage{fancyhdr}       
\usepackage{graphicx}       
\graphicspath{{media/}}     

\usepackage{xcolor}         
\usepackage{multicol}
\usepackage{multirow}
\usepackage{colortbl}

\definecolor{table_red}{rgb}{1, 0.7, 0.7}
\definecolor{table_orange}{rgb}{1,0.85, 0.7}
\definecolor{table_yellow}{rgb}{1,1, 0.8}

\title{Think Locally, Refine Globally for Memory-Efficient 3D Reconstruction}
\orglabel{}
\orglogo{%
  \makebox[\linewidth][l]{%
    \includegraphics[height=1.2cm,keepaspectratio]{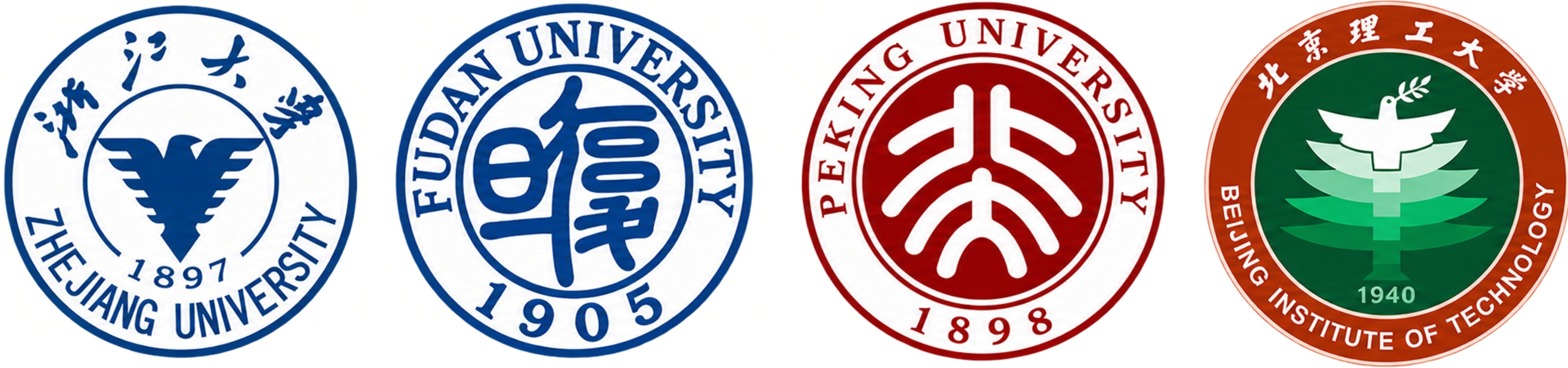}%
  }%
}

\contributors{%
  \small
  Jingke Zhou\textsuperscript{1}\quad
  Chenhang Ma\textsuperscript{1}\quad
  Zhizhou Zhong\textsuperscript{2}\quad
  Mingkai Liu\textsuperscript{3}\quad
  Zhuang Zhou\textsuperscript{4}\quad \\
  Yicheng Ji\textsuperscript{1}\quad
  Binghua Su\textsuperscript{4}\quad
  Bo Cai\textsuperscript{4} \quad
  Xianliang Huang\textsuperscript{2*,\dag}\quad
}

\contriblegend{
    \textsuperscript{1} Zhejiang University \quad
    \textsuperscript{2} Fudan University \quad
    \textsuperscript{3} Peking University  \quad 
    \textsuperscript{4} Beijing Institute of Technology \quad \\
    \textsuperscript{*}\,Corresponding author \quad 
    \textsuperscript{\dag}\,Project leader \quad 
    Contact: \href{mailto:huangxl21@m.fudan.edu.cn}{\texttt{huangxl21@m.fudan.edu.cn}}
}

\begin{document}
\maketitle
\begin{figure}[htbp]
	\begin{center}
		\includegraphics[width=1.0\linewidth]{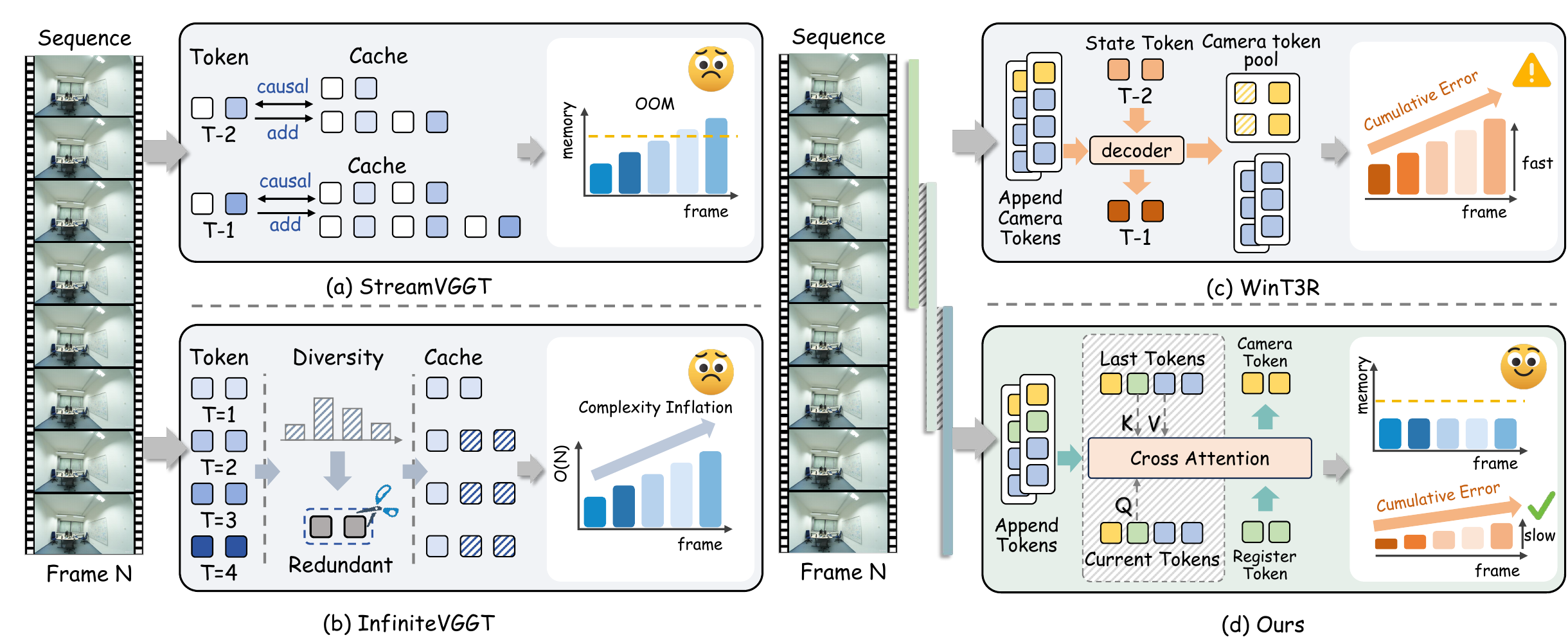}
	\end{center}
    \caption{Comparison with previous baselines on long-horizon reconstruction. Our method achieves lower accumulated error while maintaining bounded memory consumption on long sequences.}
	\label{fig:teaser1}
\vspace{-0.4cm}
\end{figure}

\begin{abstract}
We propose LoG-VGGT, a memory-efficient framework for long-sequence 3D reconstruction that balances local temporal modeling with global camera consistency.
Instead of relying on full global attention, our method introduces cross-window attention at a small subset of transformer blocks, enabling effective information propagation across adjacent temporal windows while keeping memory usage bounded.
To mitigate long-term pose drift, we further design a global camera consistency refinement module, where camera tokens interact with compact register tokens via cross-attention to enforce scene-level constraints across the entire sequence.
This design enables joint optimization of camera representations and significantly improves long-horizon pose stability without incurring the high cost of sequence-wide attention.
Extensive experiments demonstrate that LoG-VGGT achieves improved depth accuracy and robust camera pose estimation across multiple long-sequence benchmarks, while delivering competitive streaming reconstruction performance.
\end{abstract}

\section{Introduction}
\label{sec:intro}
Visual Geometry Grounded Transformer (VGGT)~\cite{wang2025vggt}, a recently proposed end-to-end reconstruction technique, has demonstrated strong performance in joint depth and pose estimation and inferring 3D attributes under offline settings. This new paradigm from 2D images is gradually serving as the bedrock for applications such as SLAM~\cite{maggio2025vggt,deng2025vggt,he2026tango3d}, 3D object editing~\cite{sun2026roar, huang2026semantic, zhou2026ov3dseg, yang2026handedit} and embodied AI~\cite{ge2025vggt,vuong2025improving,chen2025focused,zhu2026mint,yang2026bench2dex}, which previously dominated by Structure-from-Motion (SfM)~\cite{schonberger2016structure} and Multi-View Stereo(MVS)~\cite{furukawa2015multi,gu2020cascade}. 
Despite the growing impact across downstream tasks, the computational cost and memory footprint of VGGT grow prohibitively with sequence length, severely limiting their scalability to long-horizon 3D reconstruction scenarios.


This challenge arises from the quadratic complexity of global attention over spatio-temporal tokens, which becomes increasingly impractical as the sequence expands. Recent efforts to scale VGGT-style models have diverged into two distinct paradigms. Offline batch methods~\cite{deng2025vggt,wang2025vggt,shen2025fastvggt} process entire sequences jointly to maximize global consistency, but incur substantial computation and memory overhead. In contrast, 
memory-efficient approaches~\cite{lan2025stream3r,wang2025continuous,zhuo2025streaming} aim to support incremental inference by retaining all historical tokens or compressing past information into a global state. The former leads to unacceptable memory consumption, while the latter often suffers from catastrophic forgetting over long sequences, resulting in degraded geometric consistency.

Among existing long-sequence VGGT variants shown in Fig.~\ref{fig:teaser1}, StreamVGGT~\cite{zhuo2025streaming} caches all historical key value tokens in a causal transformer, enabling incremental inference but suffering from severe memory growth and out-of-memory failures on long sequences.
In Fig.~\ref{fig:teaser1}(c), WinT3R~\cite{li2025wint3r} introduces a sliding-window strategy together with a global camera token pool, allowing dense information exchange among frames within each window. Although this design successfully bounds memory consumption, it relies on maintaining a persistent global hidden state across windows.
As the sequence length increases, small local estimation errors accumulate and propagate through the global state, resulting in unstable hidden representations and significant long-term pose drift.
More recently, InfiniteVGGT~\cite{yuan2026infinitevggt} (Fig.~\ref{fig:teaser1}(b)) tackles the memory explosion issue in StreamVGGT by introducing a rolling memory mechanism with a bounded but adaptive key value cache. However, it still relies on token-level eviction heuristics that operate independently at each layer and does not explicitly enforce cross-window geometric consistency. As a result, global camera alignment is only preserved through rolling memory, which may still lead to gradual pose drift under challenging trajectories.

These limitations reveal a fundamental tension in VGGT models: \emph{can we retain sufficient temporal context for geometric consistency while preventing unbounded memory growth and long-term error drift?}
To address this challenge, we propose {LoG-VGGT}, a memory-efficient framework that \emph{thinks locally and refines globally} for scalable long-horizon 3D reconstruction.
Instead of caching all historical tokens or maintaining a fragile global state, our approach propagates information through lightweight cross-window attention at key layers, preserving local temporal continuity with bounded memory.
Furthermore, we introduce a global camera consistency refinement module, where camera tokens interact with compact register tokens, enabling global pose correction and long-term geometric consistency without reintroducing expensive full-sequence attention.

Overall, our approach strikes a favorable balance between efficiency and long-sequence modeling capacity. The main contributions are summarized as follows: (1) We introduce {LoG-VGGT}, a scalable feed-forward solution to overcome the fundamental efficiency limitations of VGGT, enabling thousands of sequences online reconstruction while preserving better geometric consistency. (2) 
We design streaming backbone with Cross-Window Attention (CWA) to enable effective inter-window information propagation under strict streaming constraints.
(3) We present a lightweight \emph{global camera consistency refinement} (GCCR) module that performs cross-attention on register tokens to jointly optimize camera tokens, effectively mitigating long-sequence pose drift. (4) Experiments on several benchmarks demonstrate that our method achieves state-of-the-art results in long-sequence reconstruction and comparable performance in streaming 3D reconstruction, while reducing computational overhead compared with existing online reconstruction methods.

\section{Related Work}
\label{sec:rela}
\subsection{Incremental 3D Reconstruction in SLAM}
Incremental 3D reconstruction has long been a core research topic in the field of simultaneous localization and mapping (SLAM). Traditional systems such as ORB-SLAM~\cite{mur2015orb} and DSO~\cite{wang2017stereo} rely on keyframe-based optimization and sliding-window bundle adjustment to incrementally refine the estimated geometry. These methods achieve accurate results, but suffer from complex optimization pipelines and limited scalability when dealing with long-sequence inputs.  
Recent advances in neural implicit representations~\cite{zhu2022nice,wang2023co} combine dense neural field optimization with real-time tracking to enable online scene reconstruction. However, these methods rely heavily on explicit geometric constraints and iterative optimization procedures, which introduce substantial computational overhead and limit scalability and speed. Moreover, they struggle to handle sparse viewpoints effectively. These challenges motivate the need for end-to-end architectures capable of learning incremental 3D reconstruction from image sequences, reducing reliance on handcrafted optimization pipelines.



\subsection{Feed-Forward 3D Reconstruction}
Feed-forward methods~\cite{zhang2025review,zhang2025advances} have recently emerged as a compelling alternative to optimization-based pipelines, offering superior speed and generalization. Pioneering approaches such as PixelNeRF~\cite{yu2021pixelnerf} and MVSNeRF~\cite{chen2021mvsnerf} learn to infer neural radiance fields directly from multi-view images through end-to-end training. Subsequent works~\cite{liu2024mvsgaussian,jiang2025anysplat} extend this idea toward explicit and efficient 3D representations~\cite{kerbl20233d}.
More recently, Visual Geometry Grounded Transformer (VGGT)~\cite{wang2025vggt} introduced an alternating-attention design that unifies camera pose estimation, depth prediction, and 3D point cloud reconstruction within a single forward pass. However, its quadratic attention complexity severely restricts scalability to long or streaming input sequences. To address this, WinT3R~\cite{li2025wint3r} extends DUSt3R~\cite{wang2024dust3r} with an incremental transformer architecture that supports frame-by-frame reconstruction, while StreamVGGT~\cite{zhuo2025streaming} and its variants~\cite{yuan2026infinitevggt} adapt VGGT with an implicit memory caching mechanism or introduces a rolling memory with token pruning to handle long sequences. Nevertheless, these approaches still struggle to maintain global coherence and suffer from information drift, as their designs primarily emphasize local temporal continuity without fully leveraging global contextual information across extended sequences. 
Consequently, achieving globally consistent and temporally stable feed-forward reconstruction over long sequences remains an open challenge.





\begin{figure*}[t]
	\begin{center}
		\includegraphics[width=0.9\linewidth]{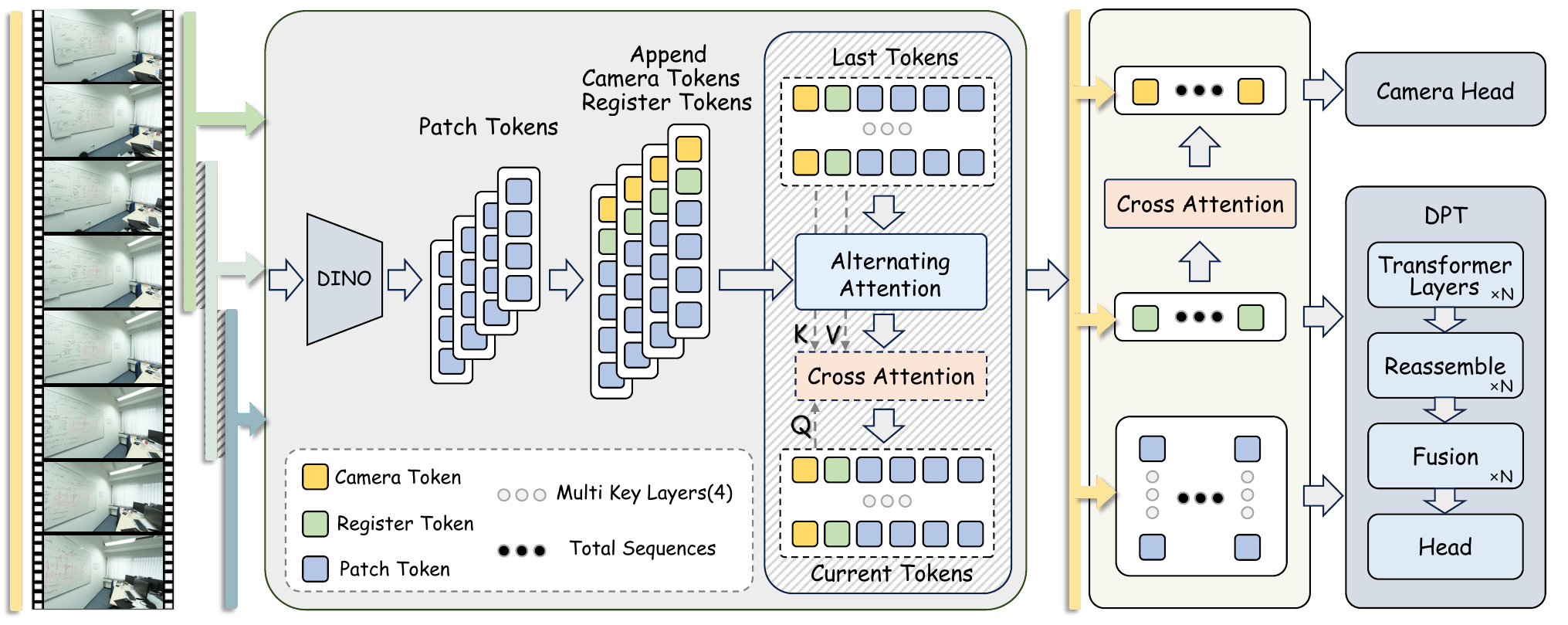}
	\end{center}
    \caption{\textbf{Overview of {LoG-VGGT}.} Given a sequences, input frames are first encoded into patch tokens using a frozen DINO visual encoder. Camera tokens and learnable register tokens are then appended to the patch tokens to form the input sequence. To efficiently model temporal dependencies under streaming constraints, we employs overlapping sliding windows, where the current tokens attend to a limited set of cached tokens from previous frames via cross-attention. Finally, a lightweight global refinement module aligns camera tokens and register tokens to mitigate long-term pose drift.}
	\label{fig:pipeline}
\vspace{-0.4cm}
\end{figure*}

\section{Motivation and Analysis}


3D reconstruction with VGGT-style models poses a fundamental challenge in keeping computation and memory bounded as the sequence grows. This motivates us to rethink how to propagate local information and refine global consistency in VGGT-style models. Existing solutions typically fall into two extremes. Full-sequence global attention maximizes consistency but incurs quadratic complexity in sequence length, making it impractical for long-horizon or streaming settings. Conversely, approaches that compress all historical information into a persistent memory state achieve bounded memory but often lose fine-grained temporal structure, leading to degraded geometric consistency over time. This motivates us to rethink how to propagate local information and refine global consistency in VGGT-style models under strict resource constraints.

One key insight is that effective long-sequence modeling does not require unrestricted global interaction at every layer. Although attention is localized within each window, the windows should not be processed in isolation, and temporal and geometric continuity must still be preserved across window boundaries. Our analysis shows that full cross-window interaction is unnecessary; instead, cross-window attention introduced at a small number of chosen transformer blocks is sufficient. Concretely, cross-window attention is applied over overlapping regions between adjacent windows, explicitly aligning shared frames to enforce local temporal and geometric consistency. Meanwhile, information is propagated gradually through these constrained pathways, implicitly inducing a weak but effective form of global consistency over long sequences. By avoiding the need to store complete token history, this design maintains constant memory complexity and strikes a favorable balance between modeling capacity and efficiency, preventing catastrophic forgetting and memory explosion.

However, while localized propagation effectively maintains mid-range geometric consistency, camera pose estimation presents a distinct challenge. Over long sequences, even minor pose errors can accumulate, resulting in significant drift when only local constraints are applied. To address this issue without reverting to expensive full-sequence attention, we introduce a lightweight global camera consistency refinement mechanism. By aggregating information from all time steps into a compact set of register tokens and performing cross-attention between these registers and camera tokens, the model can jointly refine camera poses at sequence level. 
This global refinement is independent of sliding-window attention, preserving efficiency while improving long-term pose stability.

Together, these observations motivate a design that \emph{thinks locally and refines globally}: local windowed attention ensures efficiency and short-range consistency, while global camera pose refinement corrects long-term drift. This analysis forms the foundation of our {LoG-VGGT} framework.

\section{Proposed Method}
\label{sec:method}
As illustrated in Fig.~\ref{fig:pipeline}, LoG-VGGT extends VGGT to scalable streaming 3D reconstruction under bounded memory. We first review VGGT and define the streaming reconstruction problem, then introduce the proposed memory-efficient backbone and refinement module.

\subsection{Preliminary and Problem Definition}
VGGT~\cite{wang2025vggt} is a unified feed-forward framework for geometric reasoning, capable of jointly predicting per-frame depth, camera poses, and dense 3D point representations. 
The process begins by extracting patch-level tokens from each frame using a shared vision encoder. These tokens are then processed by a stack of transformer blocks that alternate between frame attention and global attention to propagate spatial information across all frames and tokens in sequence.
Through this alternating attention scheme, VGGT produces depth maps $\hat{D}_t$, camera poses $\hat{P}_t$, and global 3D point clouds $\hat{X}_t$ in a single forward pass, achieving strong performance across a wide range of offline 3D reconstruction tasks.

Despite its effectiveness, VGGT relies on full global attention over all spatio-temporal tokens, resulting in quadratic computational and memory complexity $\mathcal{O}(N^2)$ with respect to the sequence length $N$. As the temporal horizon increases, the cost of global attention rapidly becomes prohibitive, severely limiting the scalability of VGGT for long-sequence geometric reconstruction. In this work, we address memory-efficient long-sequence 3D reconstruction under bounded computational resources. Given a potentially long image sequence $\mathcal{I}=\{I_t\}_{t=1}^{T}$, the objective is to estimate depth $\hat{D}_t$ and camera pose $\hat{P}_t$ for each frame while preserving long-range geometric consistency across the sequence.
A key challenge is to maintain sufficient cross-frame context for stable geometric reasoning without incurring unbounded memory growth or expensive sequence-wide attention.

\subsection{Memory-Efficient Streaming Backbone with Cross-Window Attention}
To enable scalable inference on image streams, we reformulate VGGT using an overlapping sliding-window backbone. Given an input sequence $\mathcal{I}=\{I_t\}_{t=1}^{T}$, we partition it into a sequence of overlapping windows:
\begin{equation}
\mathcal{W}_i = \{ I_{i}, I_{i+1}, \dots, I_{i+K-1} \},
\end{equation}
where $K$ denotes the window size and adjacent windows overlap with a fixed stride. 

Within each window, we apply the standard VGGT transformer architecture, alternating between frame attention and window-level global attention to exchange information among patch tokens, camera tokens, and register tokens. 
Crucially, global attention is restricted to tokens inside the current window, ensuring that both computational and memory complexity remain bounded by $\mathcal{O}(K^2)$, independent of the total sequence length $T$.


A direct consequence of windowed processing is the potential loss of temporal continuity across window boundaries.
Naively caching all historical tokens to address this issue is memory prohibitive.
Instead, we introduce \emph{Cross-Window Attention} (CWA) at a small subset of transformer blocks, referred to as key layers, which enables effective information propagation across windows under bounded memory.
Importantly, the key layers are not randomly selected. Following the original VGGT design, we use the same fixed set of four intermediate transformer layers that is employed for geometry prediction.
These layers have been empirically shown to encode geometry-relevant representations and serve as the primary feature sources for depth and pose estimation in VGGT.
By reusing this fixed layer set, our design remains fully consistent with the original VGGT architecture.

Let $\mathbf{H}_i^{(l)}$ denote the token representations at layer $l$ for window $\mathcal{W}_i$.
After processing window $\mathcal{W}_{i-1}$, we extract and cache the key-layer tokens $\tilde{\mathbf{H}}_{i-1}^{(l)}$ for all $l \in \mathcal{L}_{\text{key}}$.
During inference of the current window $\mathcal{W}_i$, we inject these cached tokens into the corresponding layers via cross-attention.
Formally, for each key layer $l \in \mathcal{L}_{\text{key}}$, the attention operation is defined as:
\begin{equation}
\mathbf{H}_i^{(l+1)} =
\mathcal{F}\big(
\mathbf{Q}(\mathbf{H}_i^{(l)}),\;
\mathbf{K}([\mathbf{H}_i^{(l)}, \tilde{\mathbf{H}}_{i-1}^{(l)}]),\;
\mathbf{V}([\mathbf{H}_i^{(l)}, \tilde{\mathbf{H}}_{i-1}^{(l)}])
\big),
\end{equation}
where $\mathcal{F}$ is the attention operation and $[\cdot,\cdot]$ denotes token concatenation.
The cross-window interaction explicitly aligns overlapping frames across adjacent windows, thereby enforcing local temporal continuity. By restricting cross-window attention to a fixed and small set of geometry-critical layers, our method preserves the modeling capacity of VGGT while avoiding the memory overhead associated with storing the full token history.




\subsection{Global Camera Consistency Refinement}

After streaming inference over the entire sequence, we adopt task-specific post-processing strategies for different geometric predictions.
For depth estimation, which primarily relies on local temporal continuity and progressive consistency propagation, the key-layer representations aggregated through sliding-window inference are sufficient to produce stable and coherent depth predictions.
In contrast, camera pose estimation is more susceptible to error accumulation over long temporal horizons, often resulting in noticeable pose drift.

To mitigate this issue, we introduce a \emph{Global Camera Consistency Refinement} (GCCR) module that operates at the sequence level.
Let $\mathbf{c}_t$ denote the camera token associated with frame $t$. Then, we collect register tokens from all windows and time steps, forming a global register set:
\begin{equation}
\mathbf{R} = \{ \mathbf{r}_t^j \mid t = 1,\dots,S,\; j = 1,\dots,R \},
\end{equation}
where $S$ is the number of processed windows and $R$ is the number of register tokens per window.
These register tokens preserve scene-level geometric constraints with compact global representation.
We refine each camera token by performing cross-attention with the register tokens:
\begin{equation}
\mathbf{c}_t' =
\mathrm{Attn}\big(
\mathbf{Q}(\mathbf{c}_t),\;
\mathbf{K}(\mathbf{R}),\;
\mathbf{V}(\mathbf{R})
\big).
\end{equation}

This refinement enables joint optimization of camera representations across all frames, enforcing global geometric consistency without reintroducing expensive full-sequence attention.
By decoupling local geometric reasoning from global camera alignment, the proposed module significantly improves long-horizon pose stability while maintaining the memory efficiency of the streaming framework.


\subsection{Training Objective}
Our {LoG-VGGT} is trained end-to-end in a multi-task learning framework. The overall objective combines camera pose and dense depth supervision:
\begin{equation}
\mathcal{L} = \mathcal{L}_{\text{camera}} + \mathcal{L}_{\text{depth}} ,
\end{equation}
where the two loss terms exhibit comparable magnitudes and do not require additional re-weighting.

The camera loss $\mathcal{L}_{\text{camera}}$ supervises predicted camera parameters $\hat{g}_i$, which encode translation, rotation, and intrinsics in a unified pose representation. The corresponding ground-truth parameters $g_i$ are transformed into the same encoding space. The camera loss is defined as:
\begin{equation}
\mathcal{L}_{\text{camera}} =
\frac{1}{S} \sum_{s=1}^S \gamma^{S-s}
\sum_{i=1}^N \left\| \hat{g}_i^{(s)} - g_i \right\|_1 ,
\end{equation}
where $S$ denotes the number of prediction stages and $\gamma \in (0,1]$ is a temporal decay factor that emphasizes later-stage predictions. Following VGGT~\cite{wang2025vggt}, the pose error is computed as a weighted sum of their $\ell_1$ distances:
\begin{equation}
\left\| \hat{g}_i - g_i \right\|_1 =
\lambda_t \left\| \hat{\boldsymbol{t}}_i - \boldsymbol{t}_i \right\|_1
+ \lambda_R \left\| \hat{\boldsymbol{q}}_i - \boldsymbol{q}_i \right\|_1
+ \lambda_F \left\| \hat{\boldsymbol{f}}_i - \boldsymbol{f}_i \right\|_1 .
\end{equation}

The depth loss $\mathcal{L}_{\text{depth}}$ follows DUSt3R~\cite{wang2024dust3r} and models aleatoric uncertainty. For each frame $i$, the model predicts a depth map $\hat{D}_i$ and uncertainty $\Sigma_i^D$ formulated as:
\begin{equation}
\begin{aligned}
\mathcal{L}_{\text{depth}}
=  \sum_{i=1}^N \Big(
 \big\| \Sigma_i^D \odot (\hat{D}_i - D_i) \big\|_2 
 + \big\| \Sigma_i^D \odot (\nabla \hat{D}_i - \nabla D_i) \big\|_1
- \alpha \log \Sigma_i^D \Big),
\end{aligned}
\end{equation}
where $\odot$ denotes channel-broadcast element-wise multiplication and $\alpha$ controls the regularization strength imposed on the predicted uncertainty.
Although no explicit point-map loss is used, the reconstructed 3D points are fully determined by the predicted depth and camera parameters, and are implicitly regularized by the above supervision.

\begin{table}[t]
\caption{\textbf{Quantitative 3D reconstruction results on 7-Scenes and NRGBD datasets.}}
\centering
\resizebox{\columnwidth}{!}{%
\begin{tabular}{l c | cc cc cc | cc cc cc}
\toprule
 &  & \multicolumn{6}{c|}{7-Scenes} & \multicolumn{6}{c}{NRGBD} \\
\cmidrule(lr){3-8} \cmidrule(lr){9-14}
Method & Input 
& \multicolumn{2}{c}{Acc.$\downarrow$} 
& \multicolumn{2}{c}{Comp.$\downarrow$} 
& \multicolumn{2}{c|}{NC$\uparrow$}
& \multicolumn{2}{c}{Acc.$\downarrow$} 
& \multicolumn{2}{c}{Comp.$\downarrow$} 
& \multicolumn{2}{c}{NC$\uparrow$} \\
\cmidrule(lr){3-4} \cmidrule(lr){5-6} \cmidrule(lr){7-8}
\cmidrule(lr){9-10} \cmidrule(lr){11-12} \cmidrule(lr){13-14}
 & & Mean & Med. & Mean & Med. & Mean & Med.
     & Mean & Med. & Mean & Med. & Mean & Med. \\
\midrule

CUT3R & \multirow{6}{*}{200}
& 0.132 & 0.088 & 0.065 & 0.028 & 0.547 & 0.569
& 0.221 & 0.125 & 0.068 & 0.010 & 0.583 & 0.638 \\
Point3R &
& 0.045 & 0.026 & 0.025 & 0.011 & 0.559 & 0.599
& 0.072 & 0.048 & \textbf{0.017} & 0.005 & 0.619 & 0.694 \\
TTT3R &
& 0.040 & 0.025 & 0.024 & 0.005 & 0.567 & 0.601
& 0.102 & 0.044 & 0.025 & 0.005 & 0.612 & 0.675 \\
WinT3R &
& 0.043 & 0.021 & 0.025 & 0.005 & 0.551 & 0.576
& 0.037 & 0.032 & 0.019 & 0.004 & 0.621 & 0.697 \\
InfiniteVGGT &
& 0.038 & 0.013 & 0.025 & 0.005 & 0.579 & 0.612
& 0.049 & 0.031 & 0.021 & 0.005 & 0.652 & 0.761 \\
\rowcolor{blue!8}
Ours &
& \textbf{0.036} & \textbf{0.013} & \textbf{0.024} & \textbf{0.005} & \textbf{0.609} & \textbf{0.617}
& \textbf{0.034} & \textbf{0.028} & 0.023 & \textbf{0.002} & \textbf{0.671} & \textbf{0.770} \\
\midrule

CUT3R & \multirow{6}{*}{500}
& 0.183 & 0.130 & 0.091 & 0.033 & 0.530 & 0.543
& 0.326 & 0.243 & 0.132 & 0.042 & 0.556 & 0.582 \\
Point3R &
& 0.063 & 0.026 & 0.031 & 0.015 & 0.555 & 0.583
& 0.113 & \textbf{0.048} & 0.037 & 0.005 & 0.613 & 0.684 \\
TTT3R &
& 0.062 & 0.036 & 0.029 & 0.005 & 0.552 & 0.577
& 0.165 & 0.084 & 0.095 & 0.015 & 0.594 & 0.648 \\
WinT3R &
& 0.078 & 0.040 & 0.041 & 0.015 & 0.537 & 0.553
& 0.106 & 0.065 & 0.036 & 0.006 & 0.579 & 0.621 \\
InfiniteVGGT &
& 0.043 & 0.018 & 0.025 & \textbf{0.005} & 0.561 & 0.593
& \textbf{0.080} & 0.054 & 0.037 & 0.008 & 0.643 & 0.746 \\
\rowcolor{blue!8}
Ours &
& \textbf{0.037} & \textbf{0.016} & \textbf{0.025} & 0.006 & \textbf{0.612} & \textbf{0.624}
& 0.088 & 0.057 & \textbf{0.032} & \textbf{0.004} & \textbf{0.668} & \textbf{0.759} \\
\midrule

CUT3R & \multirow{6}{*}{1000}
& 0.219 & 0.152 & 0.101 & 0.036 & 0.501 & 0.517
& 0.375 & 0.267 & 0.152 & 0.053 & 0.543 & 0.565 \\
Point3R &
& 0.082 & 0.054 & 0.049 & 0.038 & 0.545 & 0.557
& 0.150 & 0.130 & 0.151 & 0.079 & 0.588 & 0.671 \\
TTT3R &
& 0.101 & 0.077 & 0.357 & 0.096 & 0.514 & 0.533
& 0.188 & 0.095 & 0.151 & 0.053 & 0.517 & 0.565 \\
WinT3R &
& 0.134 & 0.076 & 0.055 & 0.028 & 0.522 & 0.531
& 0.197 & 0.125 & 0.124 & 0.049 & 0.548 & 0.585 \\
InfiniteVGGT &
& 0.065 & 0.038 & \textbf{0.039} & 0.027 & 0.512 & 0.535
& 0.097 & 0.089 & 0.056 & 0.028 & 0.616 & 0.698 \\
\rowcolor{blue!8}
Ours &
& \textbf{0.060} & \textbf{0.034} & 0.040 & \textbf{0.021} & \textbf{0.551} & \textbf{0.569}
& \textbf{0.091} & \textbf{0.083} & \textbf{0.051} & \textbf{0.020} & \textbf{0.647} & \textbf{0.752} \\
\bottomrule
\end{tabular}%
}
\label{tab:rec_7scenes_nrgbd}
\vspace{-0.4cm}
\end{table}

\section{Experiments}
\subsection{Experimental Settings}\label{exp:settings}


\noindent\textbf{Implementation Details.}
Our model follows the VGGT design, consisting of 24 layers of frame and global attention modules. Global attention operates at the window level with window size $K= 4$ and the stride of $2$. Training is performed in two stages: the model is first trained with batches of 12 frames for 100 epochs, followed by finetuning with batches of 64 frames for an additional 10 epochs. All experiments were conducted on a computing cluster with 16 NVIDIA H20 GPUs. More implementation details can be found in Appendix~\ref{sec:more_details}.

\noindent\textbf{Training Dataset.}
Our method is training on a curated multi-domain dataset collection consisting of 11 datasets: Co3Dv2~\cite{reizenstein2021common}, BlendMVS~\cite{yao2020blendedmvs}, ARKitScenes~\cite{dehghan2021arkitscenes}, WildRGBD~\cite{xia2024rgbd}, ScanNet~\cite{dai2017scannet}, MVS-Synth~\cite{huang2018deepmvs}, PointOdyssey~\cite{zheng2023pointodyssey}, Virtual KITTI~\cite{cabon2020virtual}, Spring~\cite{mehl2023spring},  HyperSim~\cite{roberts2021hypersim} and Replica~\cite{straub2019replica}. This collection covers diverse visual domains, spanning indoor and outdoor environments as well as varying temporal scales, with a strategic balance of data sources between synthetic generations and real-world captures. Such a hybrid composition enhances robust generalization across variations in geometric complexity, lighting conditions, and viewpoints.

\noindent\textbf{Comparison Baselines.}
We compare our method with representative streaming feed-forward 3D reconstruction baselines. Specifically, we include CUT3R~\cite{wang2025continuous}, Point3R~\cite{wu2025point3r}, and TTT3R~\cite{chen2025ttt3r} as incremental reconstruction methods, as well as VGGT-based streaming approaches WinT3R~\cite{li2025wint3r} and InfiniteVGGT~\cite{yuan2026infinitevggt}. All baselines are evaluated using their official implementations and standard configurations under comparable streaming settings. Performance is reported across 3D reconstruction, video depth estimation, and camera pose estimation tasks on multiple benchmarks.

\begin{figure*}[t]
	\begin{center}
		\includegraphics[width=0.9\linewidth]{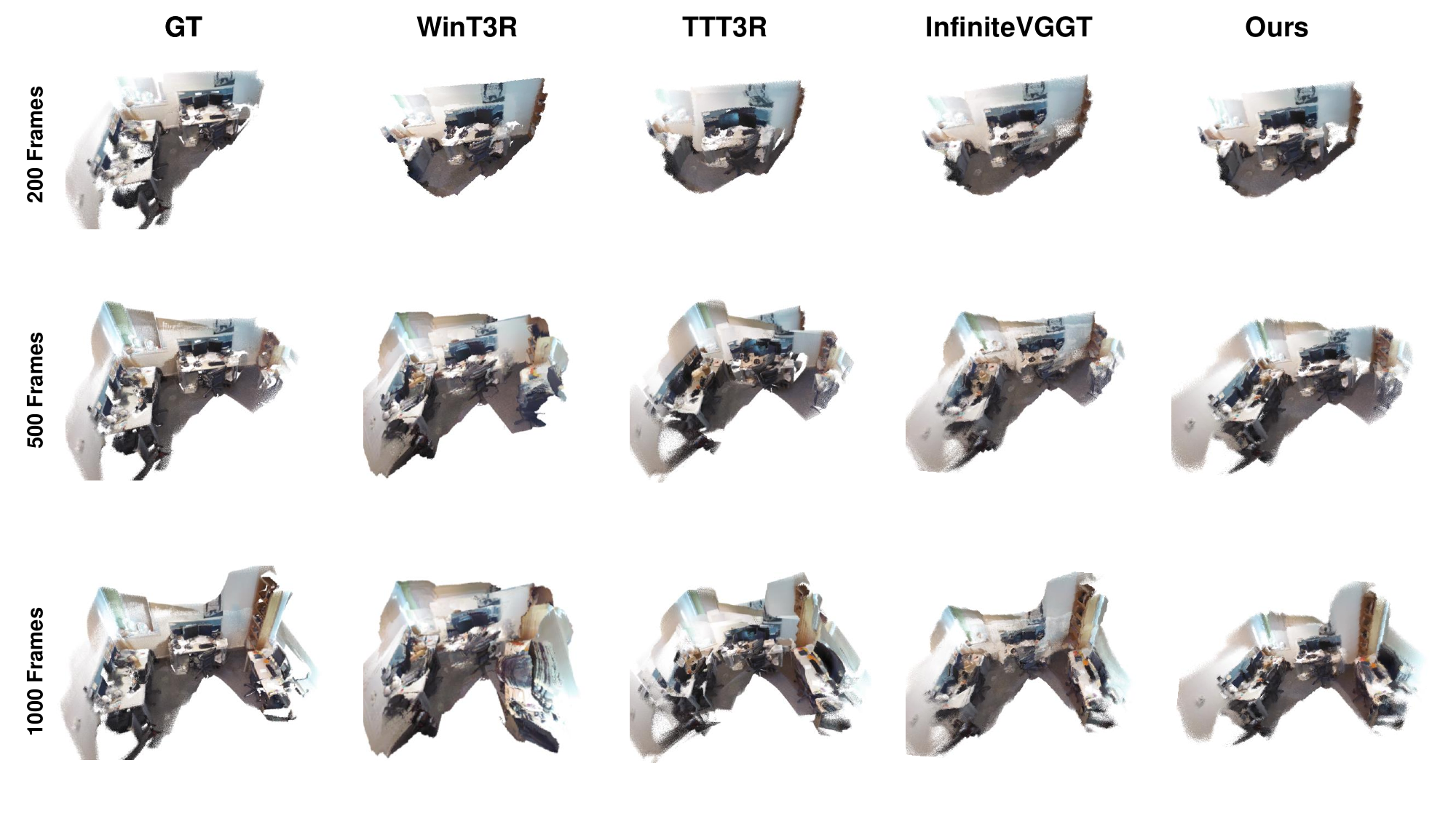}
	\end{center}
	\vspace{-0.2in}
         \caption{\textbf{Qualitative results of 3D reconstructions on 7-Scenes.} We compare 3D reconstructions from WinT3R, TTT3R, InfiniteVGGT and ours with sequence lengths increasing from top to down.}
	\label{fig:rec}
\vspace{-0.4cm}
\end{figure*}

\subsection{3D Reconstruction}
\label{exp:3D_rec}
In line with prior work, we evaluate scene-level 3D reconstruction performance on the 7-Scenes~\cite{shotton2013scene} and Neural-RGBD (NRGBD)~\cite{azinovic2022neural} datasets.
We uniformly sample subsequences of lengths 200, 500, and 1000 frames to assess performance on each sequences, with a fixed stride. We report standard metrics, including Accuracy (Acc), Completeness (Comp), and Normal Consistency (NC).

As shown in Tab.~\ref{tab:rec_7scenes_nrgbd}, our method achieves consistently strong performance across both 7-Scenes and NRGBD benchmarks under different input sequence lengths, demonstrating its robustness and effectiveness in streaming 3D reconstruction. For the 7-Scenes dataset, our approach yields the best overall accuracy and normal consistency across all tested sequence lengths, outperforming recent streaming approaches such as InfiniteVGGT and WinT3R. This advantage remains stable as the input sequence length increases to 500 and 1000 frames, indicating that our model effectively mitigates long-term drift while preserving geometric details. Fig.~\ref{fig:rec} further provides qualitative comparisons on the 7-Scenes dataset, where our reconstructions exhibit cleaner geometry, fewer artifacts, and more consistent surface normals under varying input sequence lengths. Overall, these results confirm that our method scales favorably to long sequences without sacrificing reconstruction fidelity.




\subsection{Video Depth Estimation}
\label{exp:depth_est}
We evaluate video depth estimation performance on three benchmark datasets: Virtual KITTI, Sintel~\cite{alnegheimish2022sintel}, and Bonn~\cite{palazzolo2019refusion}, using multi-frame input sequences.
These datasets collectively cover a diverse range of dynamic and static scenes across both indoor and outdoor environments.
To ensure an unbiased evaluation of cross-domain generalization, all datasets used for evaluation are strictly excluded from the training set. Following DUSt3R~\cite{wang2024dust3r}, we report two standard metrics: Absolute Relative Error (Abs Rel) and the percentage of depth predictions within a factor of 1.25 of the ground-truth depth ($\delta_{1.25}$).
As shown in Tab.~\ref{tab:video_depth}, our method achieves performance comparable to InfiniteVGGT while outperforming representative streaming-based methods across all evaluated datasets.
This suggests that our design effectively leverages multi-frame information without relying on full global attention, enabling stable depth estimation.


\begin{table}[t]
\caption{\textbf{Video depth estimation results on Bonn, KITTI, and Sintel datasets.}}
\centering
\resizebox{0.72\columnwidth}{!}{%
\setlength{\tabcolsep}{4pt}
\begin{tabular}{l | cc | cc | cc}
\toprule
\multirow{2}{*}{Method}
& \multicolumn{2}{c}{Sintel(50 frames)}
& \multicolumn{2}{c}{BONN(110 frames)}
& \multicolumn{2}{c}{KITTI(110 frames)} \\
\cmidrule(lr){2-3}
\cmidrule(lr){4-5}
\cmidrule(lr){6-7}
& Abs Rel$\downarrow$ & $\delta < 1.25\uparrow$
& Abs Rel$\downarrow$ & $\delta < 1.25\uparrow$
& Abs Rel$\downarrow$ & $\delta < 1.25\uparrow$ \\
\midrule
CUT3R &
0.421 & 0.479 & 0.078 & 0.937 & 0.118 & 0.881 \\
Point3R &
0.452 & 0.489 & 0.060 & 0.960 & 0.136 & 0.842 \\
TTT3R &
0.404 & 0.500 & 0.068 & 0.954 & 0.113 & 0.904 \\
WinT3R &
0.404 & \textbf{0.696} & 0.058 & 0.929 & 0.130 & 0.876 \\
InfiniteVGGT &
0.323 & 0.657 & 0.059 & \textbf{0.972} & 0.173 & 0.721 \\
\rowcolor{blue!8}
Ours &
\textbf{0.321} & 0.663 & \textbf{0.054} & 0.968 & \textbf{0.109} & \textbf{0.904} \\
\bottomrule
\end{tabular}%
}
\vspace{-0.4cm}
\label{tab:video_depth}
\end{table}




\begin{figure}[t]
	\begin{center}
		\includegraphics[width=0.9\linewidth]{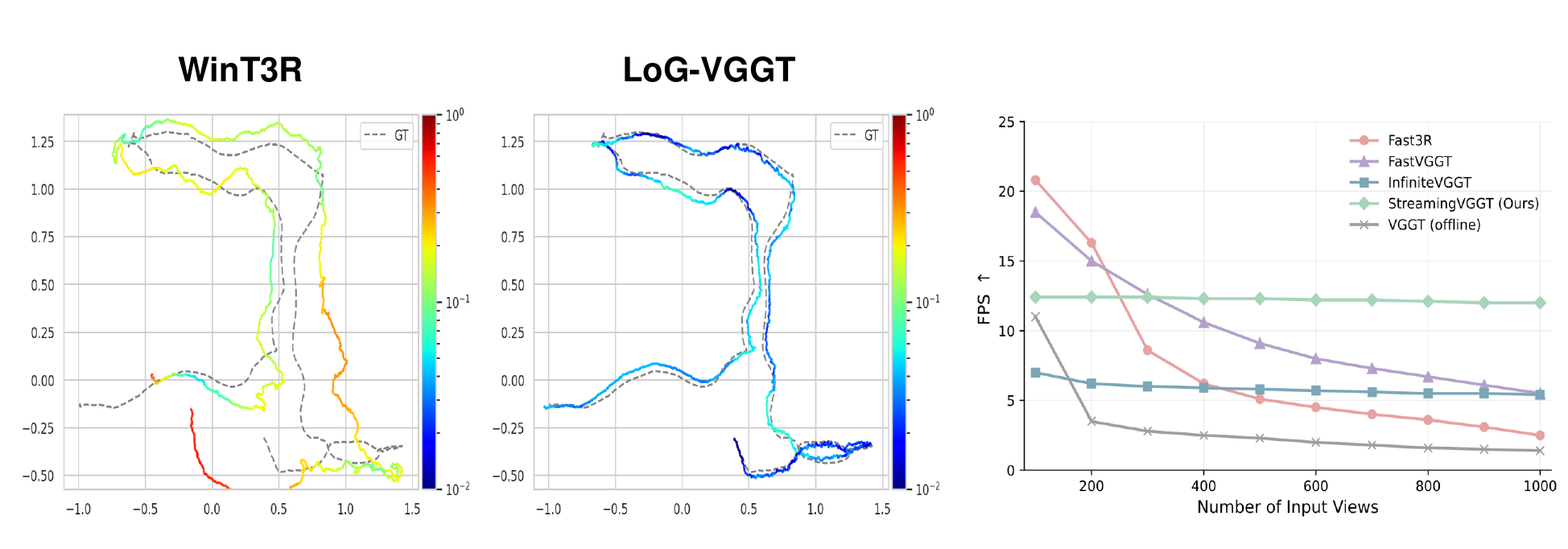}
	\end{center}
	\vspace{-0.2in}
         \caption{\textbf{Left:} Comparison of pose estimation performance between {LoG-VGGT} and WinT3R. \textbf{Right:} Inference speed comparison across different methods as the sequence length increases.}
	\label{fig:pose_est}
\vspace{-0.4cm}
\end{figure}

\subsection{Camera Pose Estimation}
\label{exp:camera_pose}
For the camera pose estimation task, we evaluate our method on ScanNet, TUM-D~\cite{sturm2012benchmark}, and CO3Dv2  datasets to ensure fair comparisons. For each scene, we randomly sample 500 frames to construct evaluation sequences.
All evaluated models are either trained on the same datasets or on none of them, avoiding potential bias from additional data.
Together, these datasets cover a broad spectrum of object-level and scene-level scenarios, as well as both static and dynamic scenarios.
We report Relative Rotation Accuracy (RRA) and Relative Translation Accuracy (RTA) at specified thresholds, e.g., RRA@30 for 30 degrees, along with AUC, a unified metric computed as the area under the accuracy–threshold curve using the minimum of RRA and RTA across thresholds.

As shown in Tab.~\ref{tab:camera_pose}, our method achieves state-of-the-art performance among streaming approaches. Furthermore, to illustrate the accumulation of pose drift commonly observed in hidden-state-based methods such as WinT3R, and to highlight the benefit of our Global Camera Consistency Refinement, Fig.~\ref{fig:pose_est} visualizes the estimated camera trajectories produced by our method and WinT3R. Our approach effectively suppresses cumulative drift over long sequences, demonstrating its robustness and practical applicability for real-world deployment.

\begin{table}[t]
\caption{\textbf{Camera Pose Estimation on ScanNet, TUM-D and CO3Dv2 datasets.}}
\centering
\resizebox{\columnwidth}{!}{%
\setlength{\tabcolsep}{4pt}
\begin{tabular}{l | ccc | ccc | ccc}
\toprule
\multirow{2}{*}{Method}
& \multicolumn{3}{c}{ScanNet}
& \multicolumn{3}{c}{TUM-D}
& \multicolumn{3}{c}{CO3Dv2} \\
\cmidrule(lr){2-4} \cmidrule(lr){5-7} \cmidrule(lr){8-10}
 & RRA@30$\uparrow$ & RTA@30$\uparrow$ & AUC@30$\uparrow$
 & RRA@30$\uparrow$ & RTA@30$\uparrow$ & AUC@30$\uparrow$
 & RRA@30$\uparrow$ & RTA@30$\uparrow$ & AUC@30$\uparrow$ \\
\midrule
CUT3R
& 58.46 & 57.13 & 54.91 & 77.62 & 73.51 & 75.59
& 76.33 & 72.67 & 75.94 \\
Point3R
& 66.73 & 62.55 & 61.06 & 78.33 & 74.12 & 76.35
& 75.51 & 71.21 & 67.99 \\
TTT3R
& 75.32 & 75.25 & 71.36 & 88.59 & 84.92 & 86.75
& 78.80 & 81.17  & 83.31 \\
WinT3R
& 61.73 & 58.25 & 56.34 & 65.60 & 68.21 & 64.66
& 73.67 & 74.52 & 75.68 \\
InfiniteVGGT
& 81.15 & 78.25 & 78.46 & 92.47 & 92.78 & 90.03
& 82.89 & 85.55 & 88.02 \\
\rowcolor{blue!8}
Ours
& \textbf{82.10} & \textbf{78.93} & \textbf{78.56}
& \textbf{92.51} & \textbf{93.46} & \textbf{91.17}
& \textbf{83.52} & \textbf{86.61} & \textbf{88.24} \\
\bottomrule
\end{tabular}%
}
\label{tab:camera_pose}
\vspace{-0.4cm}
\end{table}


\subsection{Ablation Study}\label{exp:eff}


We conduct ablation studies to evaluate the contributions of the CWA mechanism and GCCR module in {LoG-VGGT}. Experiments are performed on the 7-Scenes and NRGBD datasets, covering both 3D reconstruction and camera pose estimation. Reconstruction quality is assessed using Chamfer Distance (CD) and Normal Consistency (NC), while pose estimation is evaluated by comparing pose accuracy across ablated variants. All ablations are conducted under short-sequence (10 frames) and long-sequence (200 frames) settings to assess robustness to sequence length.


As shown in Tab.~\ref{tab:ablation}, CWA is essential for extending VGGT to the streaming setting, enabling bounded-memory processing of arbitrarily long sequences. Nevertheless, using only local cross-window interactions causes noticeable degradation on long sequences, mainly due to accumulated pose errors. 
Introducing GCCR module effectively mitigates this issue by suppressing long-term drift, leading to substantial improvements in both camera pose estimation and 3D reconstruction quality.



GCCR can be further adapted to an incremental periodic variant. We evaluate this setting on CO3Dv2 using a periodic GCCR variant that only uses past frames. As reported in Tab.~\ref{tab:gccr_comparison}, periodic GCCR already achieves clear gains over the pure streaming baseline without relying on future frames, while the full GCCR variant provides the best performance. This demonstrates an accuracy--latency trade-off, enabling users to choose between real-time responsiveness and stronger global consistency.

A detailed runtime breakdown is provided in Appendix \ref{app:runtime_breakdown} (Tab.~\ref{tab:runtime_breakdown}), showing that the streaming CWA backbone accounts for 99.84\% of the total runtime, while full-sequence GCCR contributes only 0.16\% due to its lightweight one-time cross-attention refinement over compact tokens. Appendix~\ref{app:online_gccr}(Tab.~\ref{tab:fps_comparison}) further evaluates a fully online GCCR variant, which supports frame-by-frame processing with only a minor FPS drop while maintaining a clear efficiency advantage over InfiniteVGGT.

\begin{table}[t]
\caption{\textbf{Ablation studies on 3D reconstruction and camera pose estimation.}}
\centering
\resizebox{0.72\columnwidth}{!}{%
\setlength{\tabcolsep}{4pt}
\begin{tabular}{c l | cc | cc | ccc}
\toprule
\multirow{2}{*}{Input} & \multirow{2}{*}{Method}
& \multicolumn{2}{c}{7-Scenes}
& \multicolumn{2}{c}{NRGBD}
& \multicolumn{3}{c}{CO3Dv2} \\
\cmidrule(lr){3-4} \cmidrule(lr){5-6} \cmidrule(lr){7-9}
& 
& CD$\downarrow$ & NC$\uparrow$
& CD$\downarrow$ & NC$\uparrow$
& RRA@30$\uparrow$ & RTA@30$\uparrow$ & AUC@30$\uparrow$ \\
\midrule

10 & w/o CWA
& 1.846 & 0.512 
& 2.470 & 0.495
& 67.53 & 68.54 & 42.35 \\
10 & w/o GCCR
& 0.027 & 0.634
& 0.029 & 0.669
& 74.64 & 79.27 & 62.63 \\
\rowcolor{blue!8}
10 & Full model
& \textbf{0.022} & \textbf{0.688}
& \textbf{0.021} & \textbf{0.714}
& \textbf{97.12} & \textbf{98.66} & \textbf{91.61} \\
\midrule

200 & w/o CWA
& OOM & OOM
& OOM & OOM
& OOM & OOM & OOM \\
200 & w/o GCCR
& 0.045 & 0.503
& 0.061 & 0.525
& 60.34 & 62.28 & 44.27 \\
\rowcolor{blue!8}
200 & Full model
& \textbf{0.032} & \textbf{0.609}
& \textbf{0.026} & \textbf{0.671}
& \textbf{83.52} & \textbf{86.61} & \textbf{88.24} \\

\bottomrule
\end{tabular}%
}
\label{tab:ablation}
\vspace{-0.4cm}
\end{table}


\begin{table}[h]
\centering
\caption{\textbf{Ablation studies of different GCCR settings on CO3Dv2 dataset.}}
\label{tab:gccr_comparison}
\resizebox{\linewidth}{!}{%
\begin{tabular}{lccccc}
\toprule
GCCR Setting          & Uses Future Frames & RRA@30 $\uparrow$ & RTA@30 $\uparrow$ & AUC@30 $\uparrow$ \\ \midrule
No GCCR               & $\boldsymbol{\times}$        & 60.34     & 62.28     & 44.27     \\
Periodic GCCR (every 50 frames)  & $\boldsymbol{\times}$        & 72.10     & 74.36     & 63.85     \\
Periodic GCCR (every 100 frames) & $\boldsymbol{\times}$        & 77.84     & 81.02     & 74.91     \\
\rowcolor{blue!8}
Final full-sequence GCCR         & $\boldsymbol{\checkmark}$    & \textbf{83.52}     & \textbf{86.61}     & \textbf{88.24}     \\ \bottomrule
\end{tabular}%
}
\vspace{-0.4cm}
\end{table}

\subsection{Efficiency Analysis}
\label{exp:efficiency}
To address the scalability limitations of VGGT on long sequences, InfiniteVGGT introduces rolling memory with token pruning to bound memory usage and enables infinite-horizon inference. However, this design requires maintaining and dynamically updating an explicit cache at each time step, including per-layer and per-head token selection, which incurs substantial computational overhead and leads to a degraded inference speed as the sequence length increases. In contrast, our {LoG-VGGT} enforces locality at the architectural level by restricting attention to fixed-size sliding windows. This design avoids costly cache maintenance and ensures that the computational cost per frame remains approximately constant with respect to the sequence length. As a result, LoG-VGGT achieves significantly higher and more stable inference speed across varying input lengths, as shown in Fig.~\ref{fig:pose_est}.

\section{Conclusion}
In this paper, we present {LoG-VGGT}, a memory-efficient extension of VGGT that enables scalable 3D reconstruction from streaming image sequences. 
To explicitly preserve local temporal continuity across adjacent windows, we introduce cross-window attention at a small set of geometry-critical layers while keeping memory usage bounded. Then, we propose a lightweight global camera consistency refinement module that jointly optimizes camera tokens through cross-attention with compact register tokens. This module effectively corrects long-term pose drift without incurring the cost of expensive sequence-wide attention.
Our method achieves improved depth accuracy and more stable camera pose estimation compared to existing baselines, while significantly reducing memory and computational overhead.




\bibliographystyle{unsrt}  
\bibliography{main_reference}  

\newpage
\appendix

\section{Implementation Details}\label{sec:more_details}

Our model architecture is based on the VGGT design, consisting of \(L = 24\) layers of frame and global attention modules. Global attention is applied at the window level, with a window size of 4 and a stride of 2, following the settings in WinT3R. In cross-window attention, key layers from overlapping frames in the previous window are selected and propagated as the key (\(K\)) and value (\(V\)) for attention to the next window. These key layers correspond to the following DPTHead module indices: \( \text{block4DPT\_idx} = [4, 11, 17, 23] \). These layers are crucial because the DPTHead module of VGGT relies on tokens extracted from these specific layers for prediction.

Training is performed in two stages. Initially, we train the model using batches of 12 frames for 100 epochs to stabilize the local geometric reasoning. Afterward, the model is finetuned using batches of 64 frames for an additional 10 epochs to enhance long-range temporal consistency. During each training iteration, frames are randomly sampled from diverse training sequences to ensure variability. Following the approach in Point3R~\cite{wu2025point3r}, input images with varying aspect ratios are resized such that the longer edge is set to 518 pixels, while maintaining the original aspect ratio.

All experiments are conducted on a computing cluster equipped with 16 NVIDIA H20 GPUs. This setup allows us to efficiently process large sequences and conduct extensive evaluation across multiple benchmarks.

\begin{figure}
	\begin{center}
		\includegraphics[width=\linewidth]{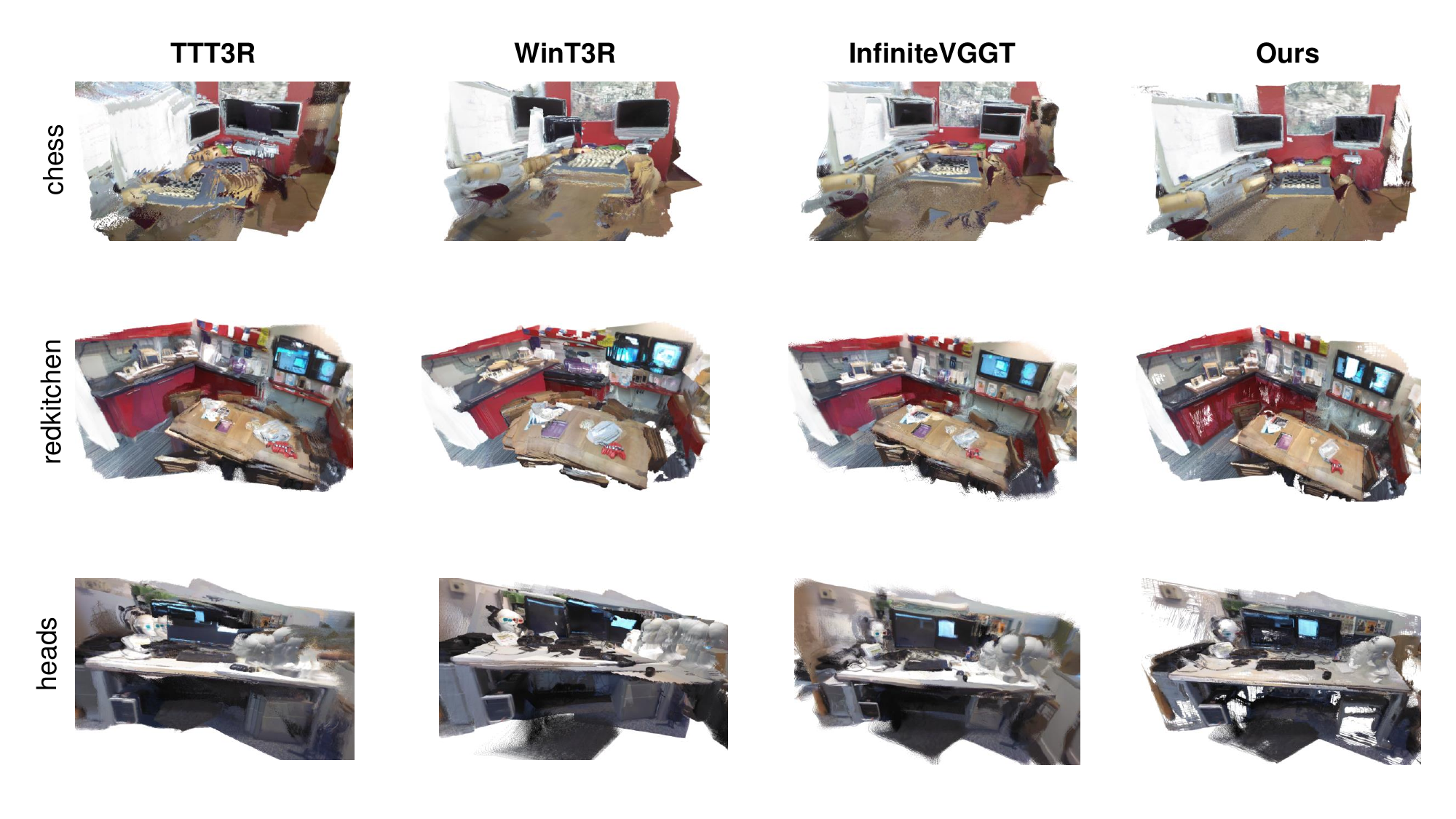}
	\end{center}
	\vspace{-0.2in}
         \caption{More Qualitative Results of 3D Reconstruction on 7-Scenes.}
	\label{fig:rec_more}
\end{figure}

\begin{figure}[t]
	\begin{center}
		\includegraphics[width=\linewidth]{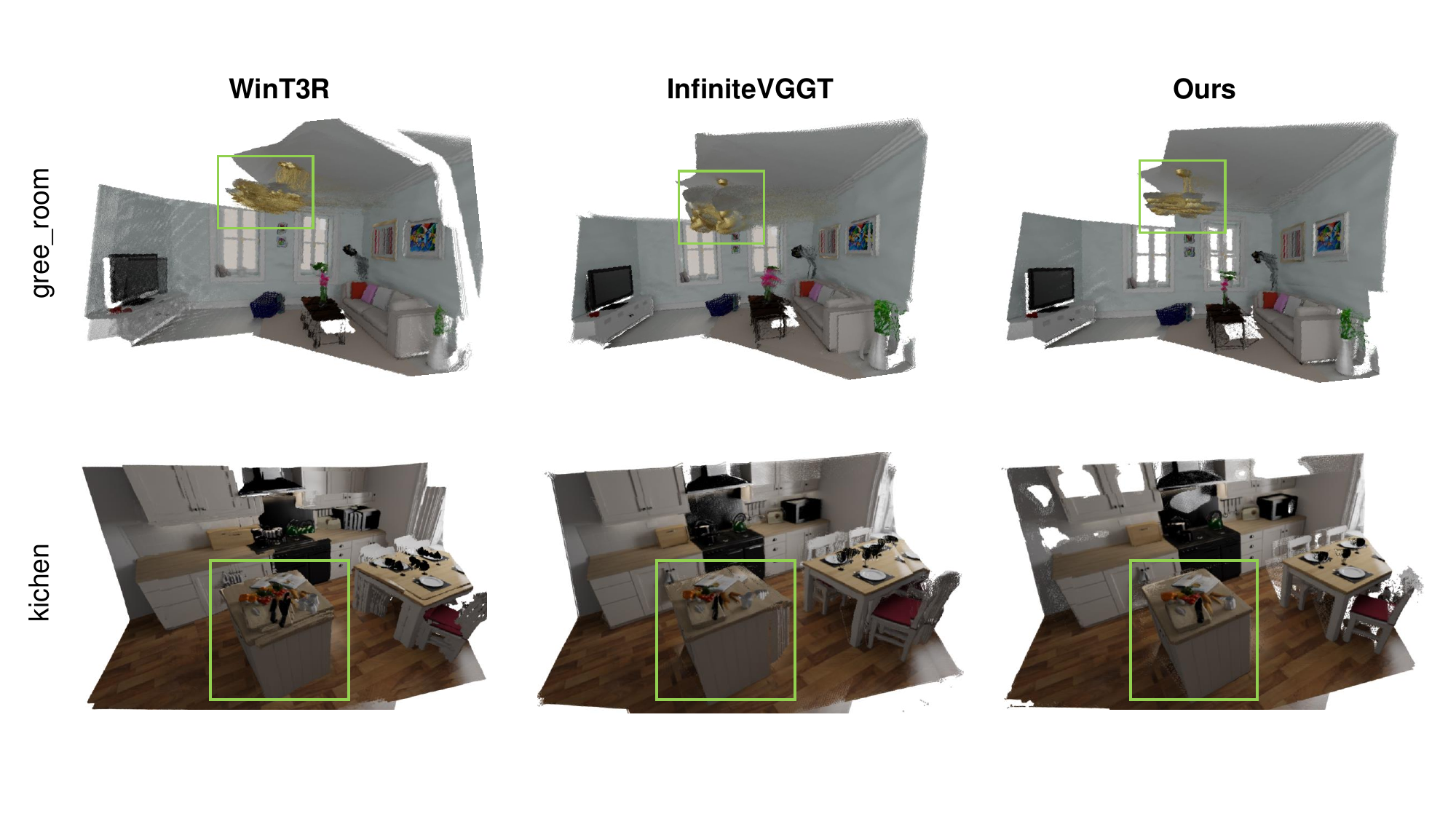}
	\end{center}
	\vspace{-0.2in}
         \caption{Qualitative Results of 3D Reconstruction on NRGBD.}
	\label{fig:rec_nrgbd}
\end{figure}

\section{Additional Results}
\label{app:more_results}
\subsection{3D Reconstruction}
Fig.~\ref{fig:rec_more} and Fig.~\ref{fig:rec_nrgbd} present additional qualitative comparisons of scene-level 3D reconstruction results between our method and other baseline approaches. On the 7-Scenes dataset, we sample each sequence with a stride of 2, resulting in approximately 500 frames per sequence for evaluation. On the NRGBD dataset, we use the first 100 frames of each sequence, following common practice for qualitative visualization. Across different scenes, our method produces more complete reconstructions with improved geometric consistency, particularly in regions affected by occlusions or limited viewpoints.

\begin{figure*}[t]
	\begin{center}
		\includegraphics[width=\linewidth]{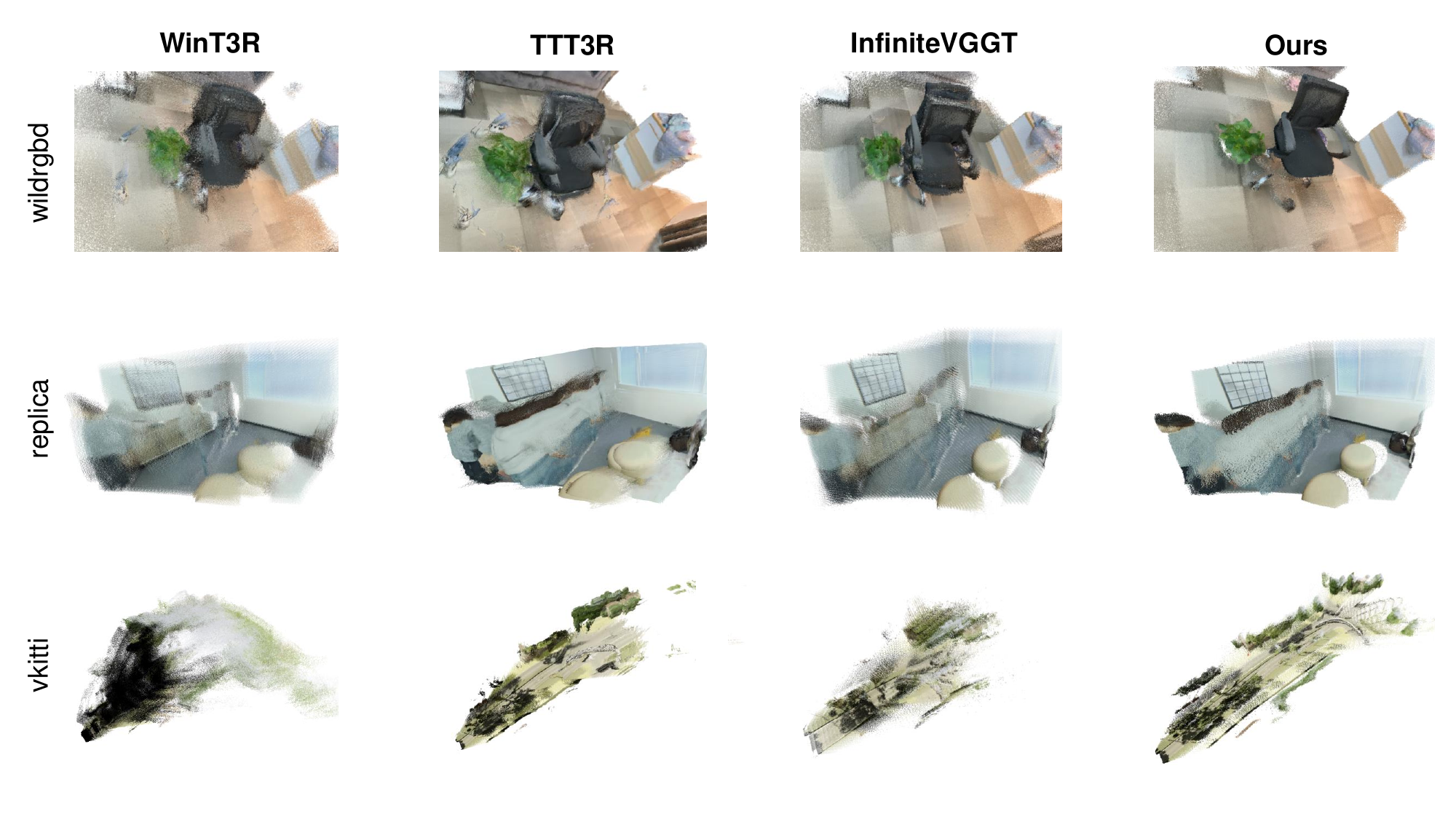}
	\end{center}
	\vspace{-0.2in}
         \caption{Qualitative Results of 3D Reconstruction on more datasets.}
	\label{fig:rec_multi}
\end{figure*}

To further evaluate the robustness of our method across diverse data regimes, Fig.~\ref{fig:rec_multi} shows qualitative results on object-centric scenes from WildRGBD, dynamic scenes from Dynamic Replica~\cite{karaev2023dynamicstereo}, and outdoor environments from Virtual KITTI.
In all settings, our method produces more complete and geometrically consistent reconstructions.

\begin{figure}[t]
	\begin{center}
		\includegraphics[width=\linewidth]{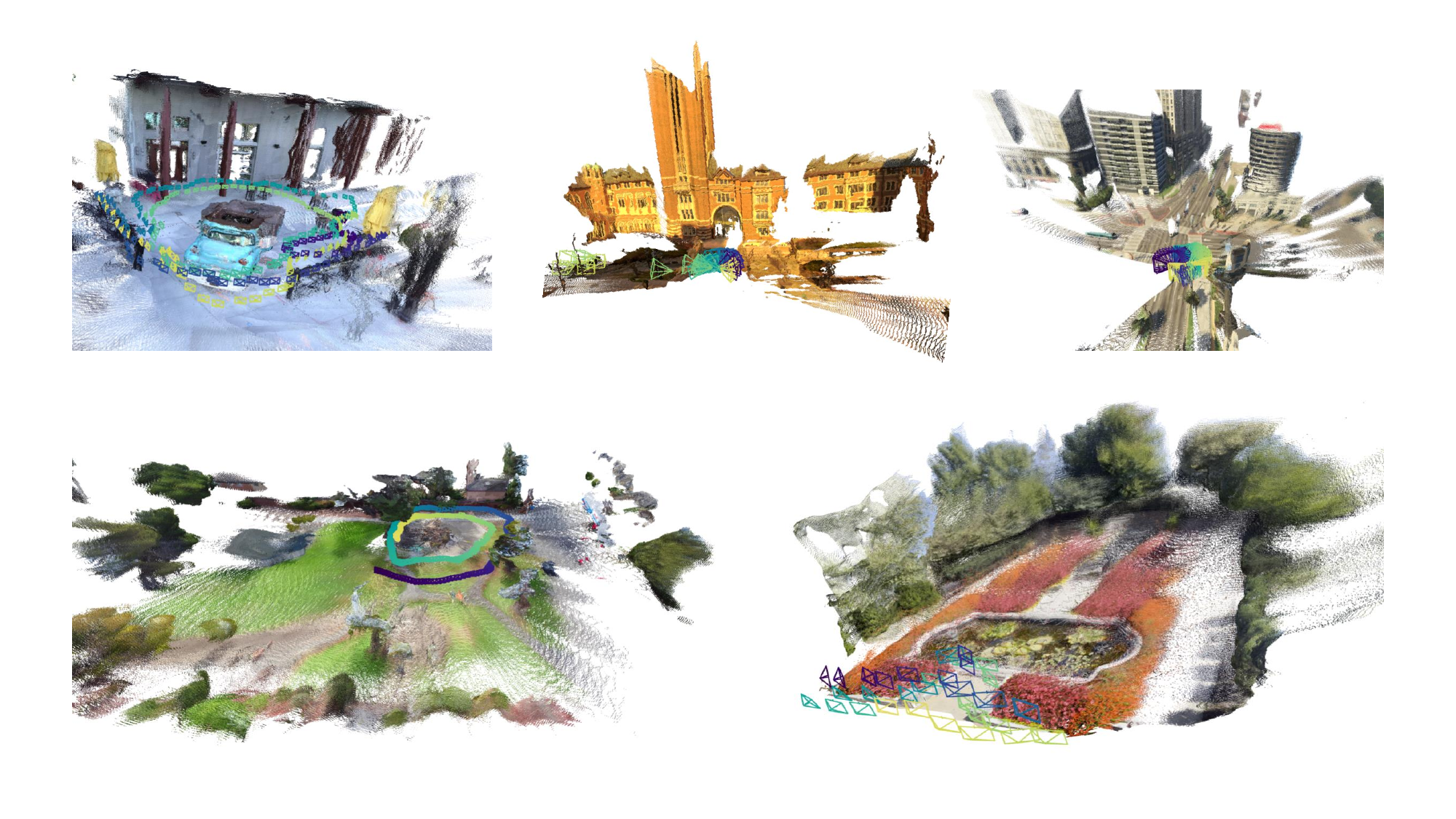}
	\end{center}
	\vspace{-0.2in}
         \caption{ Qualitative Results of 3D reconstruction in-the-wild.}
	\label{fig:rec_wild}
\end{figure}

Fig.~\ref{fig:rec_wild} shows qualitative reconstruction results of our method on large-scale outdoor long-sequence scenes, along with the estimated camera trajectories. Despite the increased scene scale and longer temporal horizon, our method is able to maintain coherent geometry and stable camera estimation over time. These results demonstrate the robustness of our approach in challenging real-world scenarios and further validate its effectiveness for long-horizon streaming 3D reconstruction.

Quantitative evaluations on the 7-Scenes and NRGBD datasets (Table~\ref{tab:perf_compare}) with XStreamVGGT as the baseline—using Accuracy (Acc), Completeness (Comp, lower is better) and Normal Consistency (NC, higher is better) under 200 and 1000 input frames—show our method’s superiority: with 200 frames, it outperforms the baseline in Acc and NC on 7-Scenes and achieves significant improvements (reduced Acc/Comp and increased NC) on NRGBD; with 1000 frames, it maintains advantages in Acc and Comp on 7-Scenes and competitive Comp on NRGBD, mitigating performance degradation from excessive frames. These quantitative results align with qualitative observations, confirming our method’s effectiveness.

\begin{table}[t]
\caption{Quantitative 3D reconstruction results on 7-Scenes and NRGBD datasets with XStreamVGGT.}
\centering
\resizebox{\linewidth}{!}{
\begin{tabular}{l l ccc ccc}
\toprule
\multirow{2}{*}{Method} & \multirow{2}{*}{Input}
    & \multicolumn{3}{c}{7-Scenes} 
    & \multicolumn{3}{c}{NRGBD} \\
\cmidrule(lr){3-5} \cmidrule(lr){6-8}
    & 
    & Acc (Mean/Med.) $\downarrow$ 
    & Comp (Mean/Med.) $\downarrow$ 
    & NC (Mean/Med.) $\uparrow$ 
    & Acc (Mean/Med.) $\downarrow$ 
    & Comp (Mean/Med.) $\downarrow$ 
    & NC (Mean/Med.) $\uparrow$ \\
\midrule
XStreamVGGT  & 200 & 0.040 / 0.022 & 0.017 / 0.003 & 0.569 / 0.606 & 0.108 / 0.095 & 0.052 / 0.023 & 0.619 / 0.701 \\
\rowcolor{blue!8}
Ours         & 200 & 0.036 / 0.013 & 0.024 / 0.005 & 0.609 / 0.617 & 0.034 / 0.028 & 0.023 / 0.002 & 0.671 / 0.770 \\
\midrule
XStreamVGGT  & 1000 & 0.142 / 0.068 & 0.125 / 0.048 & 0.734 / 0.848 & 0.085 / 0.049 & 0.075 / 0.038 & 0.850 / 0.986 \\
\rowcolor{blue!8}
Ours        & 1000 & 0.060 / 0.034 & 0.040 / 0.021 & 0.551 / 0.569 & 0.091 / 0.083 & 0.051 / 0.020 & 0.647 / 0.752 \\
\bottomrule
\end{tabular}
}
\label{tab:perf_compare}
\end{table}

\begin{table}[t]
\caption{\textbf{Video depth estimation results on Bonn and KITTI datasets.}}
\centering
\small
\setlength{\tabcolsep}{4pt}
\begin{tabular}{l c | cc | cc}
\toprule
\multirow{2}{*}{Method} & \multirow{2}{*}{Input}
& \multicolumn{2}{c}{Bonn}
& \multicolumn{2}{c}{KITTI} \\
\cmidrule(lr){3-4} 
\cmidrule(lr){5-6}
 &  & Abs Rel$\downarrow$ & $\delta < 1.25\uparrow$
      & Abs Rel$\downarrow$ & $\delta < 1.25\uparrow$ \\
\midrule

CUT3R & \multirow{5}{*}{200}
& 0.072 & 0.947 & 0.125 & 0.850 \\
Point3R  &
& 0.069 & 0.954 & 0.191 & 0.728 \\
TTT3R &
& 0.068 & 0.953 & 0.110 & 0.893 \\
InfiniteVGGT &
& \textbf{0.063} & \textbf{0.964} & 0.173 & 0.720 \\
\rowcolor{blue!8}
Ours &
& 0.065 & 0.957 & \textbf{0.110} & \textbf{0.898} \\
\midrule

CUT3R  & \multirow{5}{*}{500}
& 0.084 & 0.939 & 0.152 & 0.809 \\
Point3R  &
& 0.081 & 0.946 & 0.210 & 0.694 \\
TTT3R  &
& 0.076 & 0.953 & 0.132 & 0.868 \\
InfiniteVGGT &
& 0.069 & \textbf{0.960} & 0.179 & 0.714 \\
\rowcolor{blue!8}
Ours &
& \textbf{0.065} & 0.952 & \textbf{ 0.126} & \textbf{0.852} \\
\bottomrule
\end{tabular}
\label{tab:video_depth_multi_input}
\end{table}

\subsection{Depth Estimation}
To further evaluate depth estimation performance under extended temporal settings, we conduct additional experiments on the Virtual KITTI and Bonn datasets using longer video sequences.
Following the protocol in the main paper, we randomly sample input sequences of 200 and 500 frames from each dataset to examine the behavior of different methods as the temporal horizon increases.

All evaluation sequences are strictly excluded from the training set to ensure a fair cross-domain generalization setting.
We report the same evaluation metrics as in the main paper, including Absolute Relative Error (Abs Rel) and the percentage of depth predictions within a factor of 1.25 of the ground-truth depth ($\delta < 1.25$).

Quantitative results are summarized in Tab.~\ref{tab:video_depth_multi_input}.
As shown in the table, our method achieves competitive performance across both datasets and sequence lengths, and consistently performs favorably compared to streaming-based baselines.
In particular, our approach maintains stable accuracy as the input sequence length increases, indicating its robustness to extended temporal inputs.

\begin{figure}[t]
    \centering
    \begin{minipage}[t]{0.48\linewidth}  
        \centering
        \includegraphics[width=\linewidth]{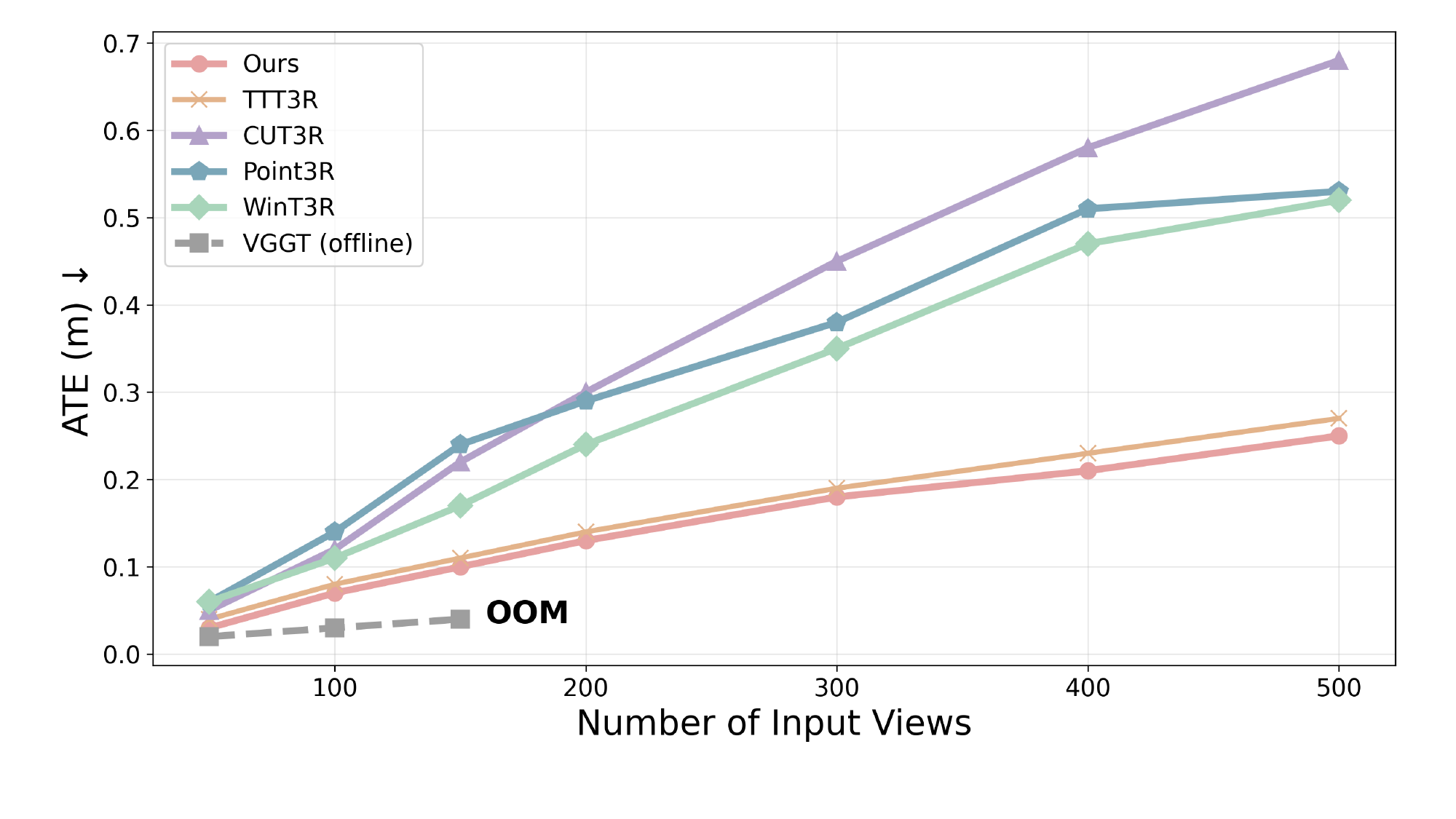}  
    \end{minipage}
    \hfill  
    \begin{minipage}[t]{0.48\linewidth}
        \centering
        \includegraphics[width=\linewidth]{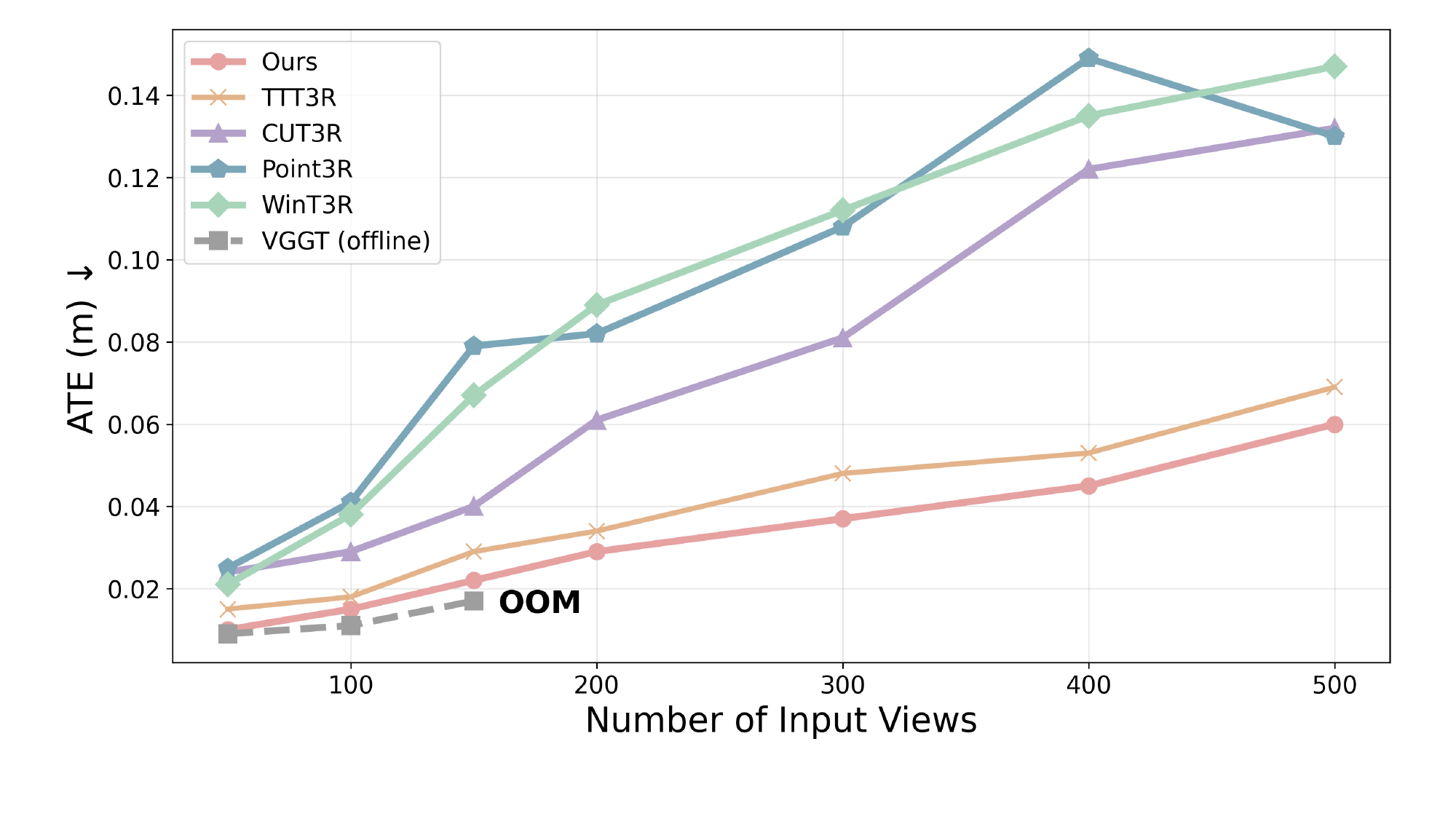}
    \end{minipage}
    
    \vspace{-0.2in}  
    \caption{Camera pose evaluation on ScanNet (left) and TUM-D (right).}  
    \label{fig:camera_pose_evaluation}  
\end{figure}

\subsection{Camera Pose Estimation} 
We further evaluate camera pose estimation performance under extended temporal horizons. Fig.~\ref{fig:camera_pose_evaluation}  report Absolute Trajectory Error (ATE) as a function of the number of input views, comparing our method with representative streaming-based baselines, including TTT3R, CUT3R, Point3R, WinT3R, and the offline VGGT model.  Compared to cache-based or hidden-state-based streaming methods, our method exhibits significantly slower error accumulation as the number of input views grows. This behavior confirms the effectiveness of the proposed Global Camera Consistency Refinement (GCCR) module in mitigating long-term pose drift without resorting to full-sequence attention. Overall, these results demonstrate that LoG-VGGT achieves a favorable balance between long-horizon pose accuracy, memory efficiency, and inference stability, making it well suited for large-scale and long-duration streaming 3D reconstruction scenarios.

\section{Additional Efficiency Analysis}

\subsection{Runtime Breakdown}
\label{app:runtime_breakdown}

Tab.~\ref{tab:runtime_breakdown} provides a detailed runtime breakdown on a 1000-frame sequence. The streaming CWA backbone accounts for 99.84\% of the total runtime, while full-sequence GCCR contributes only 0.16\%. This negligible overhead is due to the lightweight design of GCCR: instead of operating on dense 3D representations or repeatedly reconstructing global geometry, GCCR performs a one-time cross-attention refinement over compact camera and registration tokens. As a result, it effectively introduces global consistency while preserving the efficiency required for long-sequence inference.

\begin{table*}[t]
\centering
\small
\setlength{\tabcolsep}{4pt}
\renewcommand{\arraystretch}{1.08}
\begin{minipage}[t]{0.48\linewidth}
\centering
\caption{Runtime breakdown of core components on a 1000-frame sequence.}
\label{tab:runtime_breakdown}
\vspace{-2mm}
\resizebox{\linewidth}{!}{
\begin{tabular}{lcc}
\toprule
\textbf{Component} & \textbf{Runtime (ms)} & \textbf{Share} \\
\midrule
Streaming CWA backbone & 83,900 & 99.84\% \\
Full-sequence GCCR & 134 & 0.16\% \\
\midrule
\textbf{Total} & \textbf{84,034} & \textbf{100.00\%} \\
\bottomrule
\end{tabular}
}
\end{minipage}
\hfill
\begin{minipage}[t]{0.48\linewidth}
\centering
\caption{FPS comparison under different sequence lengths.}
\label{tab:fps_comparison}
\vspace{-2mm}
\resizebox{\linewidth}{!}{
\begin{tabular}{lcccc}
\toprule
\textbf{Method} & \multicolumn{4}{c}{\textbf{Sequence Length}} \\
\cmidrule(lr){2-5}
 & \textbf{100} & \textbf{200} & \textbf{500} & \textbf{1000} \\
\midrule
Fast3R & \textbf{20.8} & \textbf{16.3} & 5.1 & 2.5 \\
FastVGGT & 18.5 & 15.0 & 9.1 & 5.5 \\
InfiniteVGGT & 7.0 & 6.2 & 5.8 & 5.4 \\
\midrule
Ours w/ full-sequence GCCR & 12.4 & 12.3 & \textbf{12.3} & \textbf{11.9} \\
Ours w/ online GCCR & 11.5 & 10.8 & 9.4 & 8.2 \\
\bottomrule
\end{tabular}
}
\end{minipage}
\vspace{-2mm}
\end{table*}

\subsection{Fully Online GCCR Variant}
\label{app:online_gccr}
To support strict online scenarios, we additionally introduce a fully online GCCR variant. Unlike full-sequence GCCR, which applies a one-time global refinement after processing the entire sequence, online GCCR performs incremental cross-attention updates after each CWA window inference. This enables frame-by-frame processing without delayed batch refinement, making the method applicable to latency-sensitive streaming settings.

Tab.~\ref{tab:fps_comparison} reports the FPS comparison under different sequence lengths. Replacing full-sequence GCCR with online GCCR results in only a minor FPS drop, since the refinement still operates on compact camera and registration tokens rather than dense geometric representations. Under the strict online setting, our method maintains a clear efficiency advantage over InfiniteVGGT, confirming the effectiveness of combining streaming CWA inference with lightweight global optimization.



\section{Limitations and Broader Impacts}
\label{app:limitation}
While LoG-VGGT demonstrates strong performance and favorable scalability for long-sequence streaming 3D reconstruction, it still has several limitations. These limitations mainly arise from the central design principle of our method: introducing global reasoning only where it is most beneficial, while preserving local and streaming-friendly computation elsewhere. We discuss these limitations below, together with their broader implications and possible directions for future work.


\subsection{Limitations}
\paragraph{Depth Estimation under Local and Weak Global Constraints.}
A key design choice in LoG-VGGT is the decoupling of depth estimation from global sequence-level optimization. Motivated by the intrinsic differences between depth prediction and camera pose estimation, our method relies solely on local cross-window attention and weakly propagated temporal consistency for depth estimation, without introducing explicit global refinement across the entire sequence. This design preserves bounded memory usage and stable streaming inference, while avoiding the high cost of full-sequence attention.

Empirically, this strategy proves effective. As shown in Tab.~\ref{tab:video_depth} and Tab.~\ref{tab:video_depth_multi_input}, our method achieves performance comparable to the strongest existing streaming-based approaches, indicating that local context propagation combined with weak global consistency is sufficient for stable multi-frame depth estimation. These results support the effectiveness of our task-specific design choice.

However, our method does not consistently achieve the best performance across all depth metrics and datasets. This suggests an inherent limitation of relying exclusively on local temporal cues, particularly in scenarios involving large viewpoint changes, long-range occlusions, or globally ambiguous depth configurations. In such cases, stronger or more selective forms of global reasoning may further improve depth accuracy.

\paragraph{Efficiency Trade-off in Global Camera Refinement.}The proposed \emph{Global Camera Consistency Refinement} (GCCR) module is designed to mitigate long-term camera pose drift through compact register-token-based global reasoning. As demonstrated by the ablation results in Tab.~\ref{tab:ablation}, removing GCCR substantially degrades camera pose accuracy and reconstruction quality on long sequences. This confirms that global camera-level refinement is important for maintaining long-term geometric consistency.
Nevertheless, GCCR also introduces a small additional computational cost. Although LoG-VGGT maintains significantly more stable inference speed than cache-based methods such as InfiniteVGGT, Fig.~\ref{fig:pose_est} shows a mild decline in throughput as the sequence length increases. This is caused by the cross-attention computation between camera tokens and accumulated register tokens at the sequence level.
This overhead remains much smaller than that of full-sequence attention or dense rolling-memory mechanisms, since GCCR operates only on compact camera and register tokens rather than dense 3D representations. However, it still reflects a trade-off between long-term pose consistency and absolute inference speed. Future work may further reduce this overhead by compressing or sparsifying register tokens, performing GCCR asynchronously or at configurable temporal intervals, or designing incremental refinement schemes that reuse previously refined camera representations.

\subsection{Broader Impacts}
\label{app:broader}

This work introduces LoG-VGGT, a memory-efficient framework for long-sequence 3D reconstruction under streaming constraints. By enabling scalable geometric perception with bounded memory usage, the method may benefit applications in robotics, AR/VR, online mapping, spatial computing, and large-scale scene understanding. Its reduced computational and memory requirements may also make 3D perception more accessible on resource-constrained platforms such as mobile devices, AR/VR headsets, and small robots, while lowering inference cost and energy consumption.

At the same time, efficient long-sequence 3D reconstruction raises important privacy and security considerations. Reconstructed scenes may reveal sensitive spatial layouts, personal belongings, private indoor environments, or location-specific information. Practical deployments should therefore obtain appropriate user consent, favor on-device processing when possible, apply secure data handling, and anonymize or redact sensitive regions before storage or sharing. The technology could also be misused for unauthorized mapping or surveillance, which highlights the need for transparent deployment policies, regulatory compliance, and clear restrictions on data collection and downstream use.

In addition, performance may degrade in challenging conditions such as low-light scenes, textureless regions, reflective surfaces, fast motion, or highly dynamic environments. Before deployment in safety-critical settings, systems based on this work should be evaluated across diverse real-world scenarios and combined with uncertainty estimation or failure detection mechanisms.

Overall, LoG-VGGT contributes to scalable and efficient streaming 3D reconstruction. Its societal benefits are best realized through responsible deployment practices that prioritize privacy, transparency, safety, and robust evaluation.

\subsection{Future Directions}
Overall, these limitations point toward a broader research direction: \emph{task-adaptive global reasoning under strict streaming constraints}. Instead of applying costly global attention uniformly to all representations, future systems may selectively strengthen global reasoning for the sub-tasks that benefit most from it, such as long-term camera pose consistency, while maintaining local computation for dense predictions such as depth.
Promising directions include:\begin{itemize}  \item lightweight global depth refinement based on low-rank, hierarchical, or sparse temporal representations;  \item adaptive selection, clustering, or compression of register tokens for more efficient GCCR;  \item asynchronous or periodic global refinement to balance latency and consistency;  \item incremental refinement schemes that reuse previously refined camera representations;  \item confidence- or uncertainty-aware mechanisms that trigger global reasoning only when needed.\end{itemize}
We believe such task-adaptive designs offer a promising path toward scalable, efficient, and reliable 3D perception systems for long-sequence streaming scenarios.

\end{document}